\documentclass[11pt]{article}

\usepackage[final]{acl}

\usepackage{times}
\usepackage{latexsym}

\usepackage[T1]{fontenc}

\usepackage[utf8]{inputenc}

\usepackage{microtype}
\usepackage{inconsolata}

\usepackage{graphicx}
\usepackage{tikz}
\usepackage{pgf-pie}
\usepackage{hyperref}
\usepackage{booktabs}
\usepackage{amsmath}
\usepackage{siunitx}
\usepackage{xcolor} % For highlighting placeholders

\hypersetup{
    colorlinks=true,
    linkcolor=[RGB]{0,0,153},
    urlcolor=[RGB]{0,0,153},
    citecolor=[RGB]{0,0,153},
    linkbordercolor={0 0 0},
    pdfborder={0 0 0}
}

\title{UrduLM: A Resource-Efficient Monolingual Urdu Language Model}

\author{
  Syed Muhammad Ali Naqvi$^{*}$ \quad
  Hammad Sajid$^{*}$ \quad
  Zainab Haider$^{*}$ \quad
  Ali Muhammad Asad$^{*}$ \\
  Department of Computer Science \\
  Habib University \\
  \texttt{\{sn07590,hs07606,zh07104,aa07190\}@st.habib.edu} \\
  \AND
  Haya Fatima$^{*}$ \\
  Department of Computer Science \\
  Habib University \\
  \texttt{{sf07503}@st.habib.edu} \\
  \And
  Abdul Samad \\
  Department of Computer Science \\
  Habib University \\
  \texttt{abdul.samad@habib.edu.pk} \\
}

\begin{document}
\maketitle

\begin{abstract}
Urdu, spoken by 230 million people worldwide, lacks dedicated transformer-based language models and curated corpora. While multilingual models provide limited Urdu support, they suffer from poor performance, high computational costs, and cultural inaccuracies due to insufficient training data. To address these challenges, we present \textbf{UrduLM}, a pretrained Urdu monolingual language model trained in low-resource settings. We curate a 33GB Urdu corpus from diverse sources, develop a custom BPE tokenizer that reduces tokenization overhead by atleast 20--30\% compared to multilingual alternatives, and pretrain a 100M-parameter decoder-only model. In few-shot evaluations, UrduLM achieves competitive performance with multilingual models up to 30$\times$ its size, reaching 66.6\% accuracy on sentiment classification and BLEU scores exceeding 30 on grammar correction tasks. The complete methodology---including corpus, tokenizer, model weights, and evaluation benchmarks---is released openly to establish a baseline for Urdu NLP research and provide a scalable framework for other underrepresented languages.
\end{abstract}

\section{Introduction}

Contemporary language models remain English-centric (LLaMA-2: 89.70\% English~\cite{wendler2024llamasworkenglishlatent}), causing hallucinations and code-switching for low-resource languages~\cite{chang2023multilinguality}. The ``curse of multilinguality'' degrades performance as models support hundreds of languages, resulting in suboptimal tokenization and diluted linguistic knowledge~\cite{zhao2024largelanguagemodelshandle}.

Urdu -- being the 10th most spoken~\cite{9529190} with 230M speakers worlwide -- exemplifies these challenges. Despite widespread use, Urdu lacks dedicated transformer-based generative models, curated corpora, and standardized benchmarks. The hundreds-of-billions parameter scale of multilingual models renders them prohibitively expensive for low-resource regions like Pakistan~\cite{bender2021stochastic}, while their proprietary nature imposes data sovereignty risks. Unlike other languages with successful monolingual models (CroissantLLM~\cite{Faysse2024CroissantLLMAT}, VBART~\cite{turker2024vbartturkishllm}, PersianMind~\cite{Rostami2024PersianMindAC}), no comparable monolingual pretrained model exists for Urdu.

We introduce a systematic methodology for developing efficient monolingual LLMs in low-resource settings. Using Urdu as a case study, we demonstrate that a well-designed 100M-parameter model achieves competitive performance with multilingual models 30$\times$ larger (66.6\% sentiment accuracy, 30.59 BLEU on grammar correction), while requiring lower training cost and $\sim$32kg CO$_2$. Our complete pipeline---corpus curation, tokenizer development, training, and evaluation---provides a reproducible framework for other underrepresented languages.

\noindent \textbf{Contributions}: (1) 33GB curated Urdu corpus ($\sim$5--6B tokens) with systematic multi-source pipeline; (2) Custom BPE tokenizer achieving 20--30\% lower fertility vs GPT-4's o200k\_base; (3) 100M-parameter baseline trained for <\$100, competitive with 3B multilingual models; (4) New benchmark datasets and dual evaluation framework (LLM-as-a-Judge + few-shot).

\section{Related Work}

NLP has evolved from statistical models (HMMs, CRFs, SVMs~\cite{Khan_Daud_Nasir_Amjad_2016}) through neural embeddings (Word2Vec~\cite{mikolov2013efficientestimationwordrepresentations}, GloVe~\cite{pennington-etal-2014-glove}) and RNNs/LSTMs~\cite{LSTM} to transformer architectures~\cite{vaswani2023attentionneed}. Models like BERT~\cite{devlin-etal-2019-bert}, GPT-3~\cite{brown2020gpt3}, and T5~\cite{JMLR:v21:20-074} established \textit{foundation models}---large-scale, self-supervised models adaptable to diverse tasks. However, most are developed primarily for English.

Urdu (230M speakers~\cite{9529190}) remains severely underrepresented. Multilingual LLMs like LLaMA-2 (89.70\% English~\cite{wendler2024llamasworkenglishlatent}) exhibit diminished Urdu performance. Multilingual workflows often pivot to English for reasoning~\cite{zhao2024largelanguagemodelshandle}, losing linguistic nuance. The ``curse of multilinguality'' degrades both high- and low-resource language performance, whereas monolingual models achieve better results~\cite{chang2023multilinguality}. While parameter-efficient methods (LoRA~\cite{khade2024challengesadaptingmultilingualllms}) offer partial mitigation, they don't address inherent multilingual architectural biases.

Recent monolingual LLMs demonstrate strong performance: \textbf{CroissantLLM}~\cite{Faysse2024CroissantLLMAT} (French-English, 1.3B) achieved 87\% factual accuracy with custom bilingual tokenizer and 3T balanced tokens. \textbf{VBART}~\cite{turker2024vbartturkishllm} (Turkish, 387M-740M) reduced tokenization overhead 96\% with custom SentencePiece tokenizer, setting SOTA on Turkish benchmarks. \textbf{PersianMind}~\cite{Rostami2024PersianMindAC} shows similar benefits for Persian. These results suggest dedicated Urdu LLMs could yield comparable improvements.

\textbf{Tokenization} is critical for morphologically rich languages~\cite{jurafsky}. Standard methods (BPE, SentencePiece~\cite{sennrich-etal-2016-neural, kudo-richardson-2018-sentencepiece}) optimized for English exhibit high fertility and poor semantic splits for Urdu~\cite{rust-etal-2021-good}. Custom tokenizers trained on monolingual corpora significantly improve efficiency~\cite{Faysse2024CroissantLLMAT, turker2024vbartturkishllm, Rostami2024PersianMindAC}, yet Urdu-specific tokenization research remains lacking.

\textbf{Low-resource LLM methodologies} require addressing data scarcity, tokenization inefficiency, and evaluation challenges. Key findings: (1) Monolingual models with <1B tokens can outperform larger multilingual models when avoiding cross-linguistic interference~\cite{chang2023multilinguality}; (2) Data augmentation via parallel corpora and back-translation aids low-resource scenarios~\cite{9529190}, though at high computational cost; (3) Vocabulary extension with continued pretraining improves multilingual models~\cite{rust-etal-2021-good}, but dedicated tokenizers avoid dilution entirely; (4) Combining automated metrics with LLM-based judgments provides robust evaluation for low-resource languages~\cite{chang2023multilinguality}. Our work synthesizes these insights into a complete Urdu pipeline, transferable to other underrepresented languages.

\section{Model Architecture}

We adopt a decoder-only transformer (GPT-3 style~\cite{brown2020gpt3}) with stacked decoder blocks, multi-head self-attention, layer normalization, and feed-forward networks, enabling capture of long-range dependencies and contextual relationships.

Training leverages PyTorch's Distributed Data Parallel (DDP) and Fully-Sharded Data Parallel (FSDP) for hybrid parallelism on 4$\times$ NVIDIA V100 GPUs (32GB each). For the 100M model, DDP distributes model replicas across GPUs with gradient accumulation to simulate 0.5M token batches. Decoder-only architecture was chosen because it naturally supports generative tasks (chatbots, creative writing), requires less memory per parameter than encoder-decoder variants, and benefits from extensive GPT-series prior work providing validated hyperparameters for low-resource settings.

\section{Urdu Tokenizer Development and Evaluation}

Multilingual tokenizers exhibit poor Urdu efficiency, requiring 2--3$\times$ more tokens per word due to: (1) vocabulary dominated by high-resource languages, leaving few Urdu subword units; (2) Latin-optimized regex patterns failing on Perso-Arabic script; and (3) morphological richness causing excessive fragmentation. 

We developed a custom BPE tokenizer with Urdu-specific regex patterns and three design principles: (1) Script-Aware Segmentation: handling Urdu characters, Urdu numerals, and punctuation; (2) Optimal Vocabulary Size: exploring 10k, 20k, 32k to balance efficiency and coverage; (3) Morphological Alignment: training exclusively on Urdu text to capture language-specific patterns.

\textbf{Training Process}: We adapted OpenAI's GPT-4 regex rules for Urdu Unicode ranges (U+0600--U+06FF), applied BPE merge operations for each vocabulary size, and evaluated using fertility (tokens/word), tokenization speed, and vocabulary coverage metrics.

\begin{figure}[!ht]
  \centering
  \includegraphics[width=0.45\textwidth]{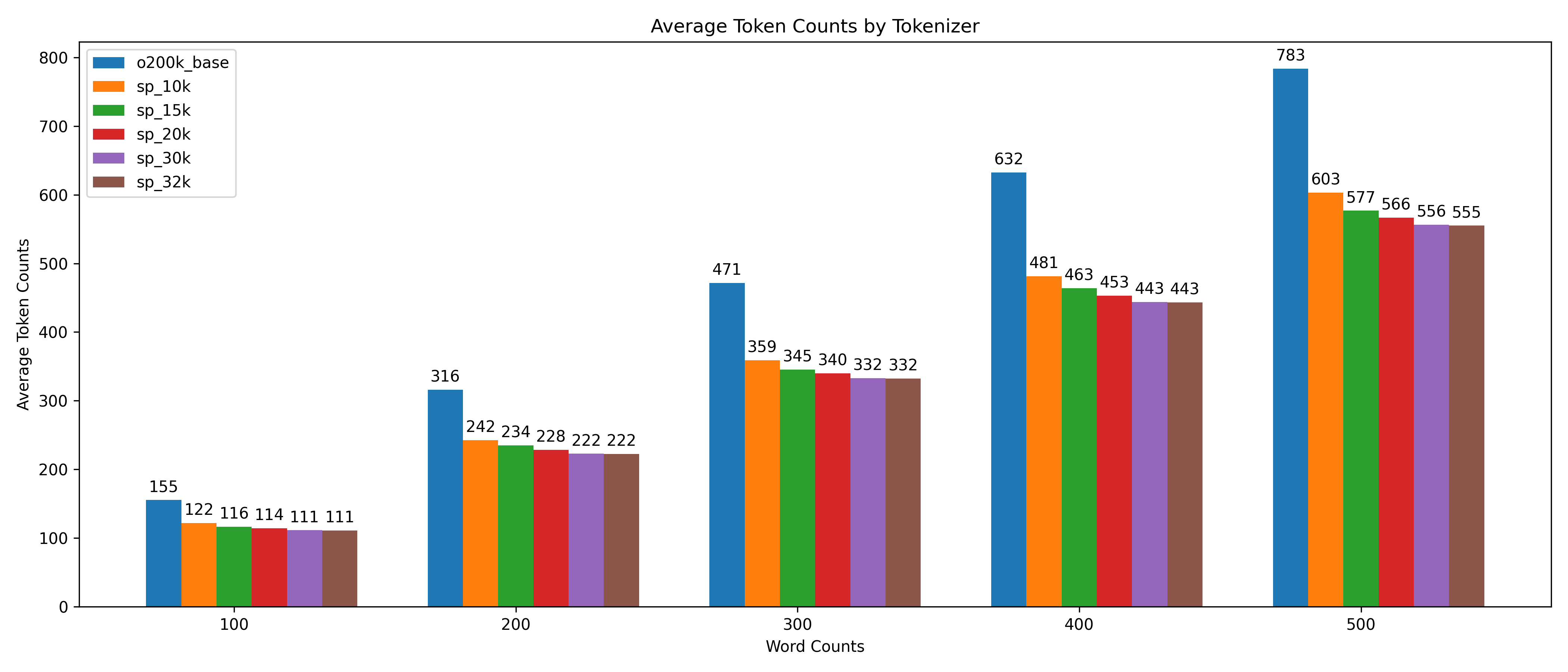}
  \caption{Tokenizer efficiency comparison: Our custom tokenizers produce 20--30\% fewer tokens than GPT-4 o200k\_base across all vocabulary sizes.}
  \label{fig:average-token-counts}
\end{figure}

Figure~\ref{fig:average-token-counts} shows our tokenizers consistently produce 20--30\% fewer tokens than o200k\_base. The 32k variant achieved best results: low fertility (comparable to 30k), broader vocabulary coverage reducing out-of-vocabulary issues, and reasonable embedding size (32k $\times$ 768 = 24.6M parameters). Accordingly, we adopted the 32k tokenizer for our final model.

\begin{table}[h]
\centering
\small
\begin{tabular}{lcc}
\toprule
\textbf{Tokenizer} & \textbf{Fertility} & \textbf{Avg Token Count}  \\
\midrule
o200k\_base & {1.566} & {783} \\
UrduLM-10k & {1.206} & {603} \\
UrduLM-32k & {1.11} & {555} \\
\bottomrule
\end{tabular}
\caption{{Fertility, Average Token Count, and Speed for o200k\_base, and our UrduLM tokenizers (500 word paragraph)}}
\label{tab:tokenizer-metrics}
\end{table}

\section{Methodology: Data Curation and Pre-Training}

\subsection{Pre-Training Dataset}

\begin{figure}[h]
  \centering
  \includegraphics[width=\linewidth, height=0.4\textheight, keepaspectratio]{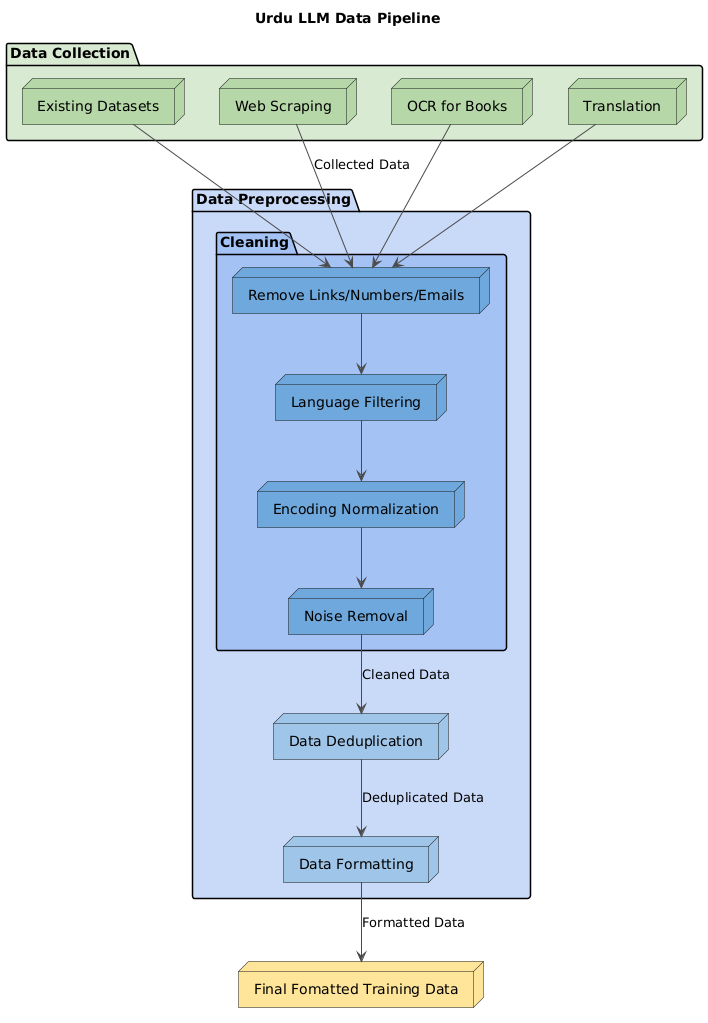}
  \caption{Overview of the data collection and preprocessing pipeline for the UrduLM pre-training dataset.}
  \label{fig:data_pip}
\end{figure}

We construct a large-scale Urdu pre-training dataset entirely from scratch, addressing the scarcity of publicly available resources for this low-resource language. Our pipeline (illustrated in Figure~\ref{fig:data_pip}) aggregates text from diverse domains including news, literature, education, and informal conversations, resulting in 33 GB of cleaned, high-quality Urdu text. This dataset, to our knowledge one of the largest curated for Urdu, provides a robust foundation for pre-training UrduLM and advancing research in Urdu language modeling.

\subsubsection{Data Collection}

To capture the richness and real-world usage of Urdu, we aggregate text from \textbf{diverse, high-coverage sources} spanning multiple domains. Our objective is to balance formal and informal registers while maximizing linguistic diversity. The dataset draws from four primary categories:

\begin{itemize}
    \item \textbf{Existing Resources}: Publicly available Urdu datasets, including UrduHack~\cite{urduhacklib} and open-source corpora such as FineWeb2~\cite{penedo2025fineweb2pipelinescale}, are incorporated as an initial seed for the collection.
    
    \item \textbf{Web Scraping}: Large-scale web crawling extracts Urdu content from blogs, forums, news sites, poetry platforms, and Urdu Wikipedia. Domain balancing is applied to prevent over-representation of any single source.
    
    \item \textbf{OCR of Books}: To enrich literary content, scanned Urdu books are digitized using the Google Cloud Vision Optical Character Recognition (OCR) API, followed by manual verification and post-processing to correct extraction errors. The API is a proprietary service used under the \href{https://cloud.google.com/terms}{Google Cloud Terms of Service}. No copyrighted or personal content is redistributed.

    \item \textbf{Translated Content}: Selected portions of the English FineWeb~\cite{penedo2024the} dataset are translated into Urdu using the Google Cloud Translation API, injecting structured educational content often missing from naturally occurring Urdu data. The Translation API is a proprietary service used under the \href{https://cloud.google.com/terms}{Google Cloud Terms of Service}, and outputs are used solely for research purposes.
\end{itemize}

This \textbf{multi-source strategy} yields a broadly representative corpus, blending formal news text, literary books, informal online conversations, and educational material. The collected raw data forms the basis for subsequent cleaning, normalization, and deduplication. All collected and processed data are intended solely for research and educational use, consistent with the licenses and access conditions of their original sources.

\subsubsection{Cleaning}

To prepare the collected corpus for downstream tasks, we developed a robust preprocessing pipeline tailored to the linguistic complexities of Urdu. Inspired by established frameworks such as ParsBERT~\cite{Farahani_2021}, the cleaning process combined regex-based noise removal with language-specific normalization techniques to produce a standardized and high-quality dataset.

\paragraph{Key components of the pipeline include:}

\begin{itemize}
    \item \textbf{Regex-Based Noise Removal:} Custom regular expressions removed non-linguistic content such as URLs, emails, phone numbers, stray digits, Latin script characters, and extraneous symbols. An optional flag (\texttt{remove\_english}) allowed conditional removal of embedded English text.
    
    \item \textbf{Digit Conversion:} All English numerals (0--9) were mapped to their Urdu counterparts to maintain script consistency throughout the dataset.
    
    \item \textbf{Character Normalization:} We used an augmented character mapping (based on UrduHack's~\cite{urduhacklib} correction dictionary) to normalize commonly confused or erroneous Urdu characters, aligning with ParsBERT-style normalization practices~\cite{Farahani_2021}.
    
    \item \textbf{Word Spacing Correction:} Incorrectly concatenated or misjoined words were segmented using UrduHack's~\cite{urduhacklib} word-space mapping, ensuring orthographic accuracy and proper word boundaries.
    
    \item \textbf{Unicode and Symbol Cleanup:} Additional post-processing included collapsing multiple whitespaces, removing invisible Unicode characters (\texttt{\textbackslash u00A0}, \texttt{\textbackslash u200B}), eliminating redundant punctuation (e.g., repeated Urdu question marks), and discarding empty parentheses.
\end{itemize}

This systematic pipeline produced clean and linguistically sound Urdu text, optimized for subsequent deduplication and tokenization steps.

\subsubsection{Deduplication}

To ensure dataset quality and integrity, we apply a rigorous deduplication process aimed at removing redundant content both within and across data sources. Redundancy not only inflates dataset size but can bias language models during pretraining, reducing generalization and efficiency.

\paragraph{Our deduplication pipeline involves three key stages:}

\begin{itemize}
    \item \textbf{Near-Duplicate Detection via LSH:} We adopt a scalable MinHash-based Locality Sensitive Hashing (LSH) approach to identify near-duplicate text segments. Each document is encoded into a hash signature, and LSH bucketing is applied to group similar documents for comparison.

    \item \textbf{Duplicate Removal:} Within each LSH bucket, exact duplicate entries are removed. In cases of high similarity (>90\% Jaccard similarity), only the most representative document is retained to preserve diversity while eliminating redundancy.

    \item \textbf{Manual Validation:} Following automated deduplication, a manual pass is conducted to ensure the dataset is free from residual duplicates or inconsistencies. This final step guarantees the retention of only unique and high-quality text instances.
\end{itemize}

This deduplication framework significantly reduces noise, improves storage efficiency, and enhances the reliability of the dataset for downstream tasks such as model pretraining and evaluation.

\subsubsection{Dataset Formatting and Final Statistics}

The processed corpus is stored in a structured CSV format with three columns: 
\texttt{data} (extracted text), \texttt{source} (origin, e.g., specific news or book site), and 
\texttt{category} (semantic domain such as news, blogs, books). During pretraining, 
samples are separated by explicit End-of-Text (EOT) tokens, allowing the model to 
learn document boundaries and handle heterogeneous text segments effectively.

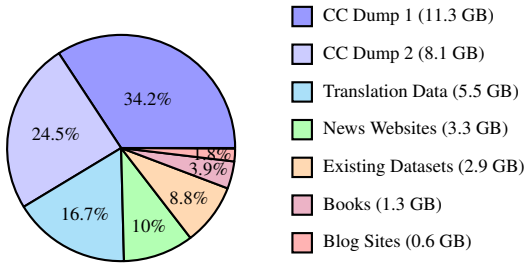
\begin{figure}[t]
    \centering
    \begin{tikzpicture}
        \pie[
            text=legend,
            radius=1.5,
            before number=\scriptsize,
            after number=\%, 
            color={blue!40, blue!20, cyan!30, green!30, orange!30, purple!30, red!30},
            font=\scriptsize
        ]{
            34.2/CC Dump 1 (11.3 GB),
            24.5/CC Dump 2 (8.1 GB),
            16.7/Translation Data (5.5 GB),
            10/News Websites (3.3 GB),
            8.8/Existing Datasets (2.9 GB),
            3.9/Books (1.3 GB),
            1.8/Blog Sites (0.6 GB)
        }
    \end{tikzpicture}
    \caption{Distribution of data sources in the UrduLM pretraining dataset.}
    \label{fig:dataset-pie}
\end{figure}

Following the pipeline, the final dataset comprises \textbf{33\,GB of high-quality Urdu text}, totaling approximately \textbf{13 million rows} and an estimated \textbf{5--6 billion tokens}. As this corpus is used exclusively for unsupervised pretraining, no train/validation/test splits are defined. This broad coverage across domains and writing styles ensures a rich training signal for UrduLM. The distribution of sources is illustrated in Figure~\ref{fig:dataset-pie}. A preview of the dataset has been made available to the public, and can be viewed \href{https://huggingface.co/datasets/orature/ALIF_Urdu_Corpus_AUC}{here}.

\paragraph{Licensing and Usage} 
The FineWeb2 dataset~\cite{penedo2025fineweb2pipelinescale} is released under the \textit{Open Data Commons Attribution License (ODC-By v1.0)} and its associated processing code under the \textit{Apache 2.0} license.
The UrduHack~\cite{urduhacklib} toolkit, used for text normalization, includes components licensed under \textit{MIT} (e.g., the \texttt{urdu-characters} submodule).
All data usage aligns with the original terms of use and is limited strictly to research purposes.

\subsection{Model Pre-training Configuration}

UrduLM-100M is a compact 100 million parameter base model designed for low-resource environments. To explore the impact of vocabulary size on model performance, we trained three variants of the 100M model using tokenizers with vocabulary sizes of 10k, 20k, and 32k tokens, respectively. This allowed us to analyze how different tokenizer configurations affect model loss and performance. Among these, the model utilizing the 32k vocabulary tokenizer demonstrated the best overall results, which are discussed in Section~\ref{sec:evaluation}. 

Detailed configurations and training setups for each variant are summarized in Table~\ref{tab:model-configs}. Here $n_{\text{params}}$ is the total number of trainable parameters, $n_{\text{layers}}$ is the total number of layers, $d_{\text{model}}$ is the number of units in each bottleneck layer, $d_{\text{head}}$ is the dimension of each attention head, $n_{\text{ctx}}$ is the context length of the model, and Vocab Size is the number of rows of the embedding table.

\begin{table*}[h]
\centering
\small
\setlength{\tabcolsep}{4pt}
\begin{tabular}{lccccccccc}
\toprule
\textbf{Model} & {$n_{\text{params}}$ (M)} & {$n_{\text{layers}}$} & {$d_{\text{model}}$} & {$n_{\text{heads}}$} & {$d_{\text{head}}$} & {$n_{\text{ctx}}$} & \textbf{Batch (M)} & \textbf{Vocab (K)} & \textbf{LR} \\
\midrule
UrduLM-100M-10k & 100 & 12 & 768 & 12 & 64 & 1024 & 0.5 & 10 & 6.0e-4 \\
UrduLM-100M-20k & 116 & 12 & 768 & 16 & 64 & 1024 & 0.5 & 20 & 6.0e-4 \\
UrduLM-100M-32k & 134 & 12 & 768 & 16 & 64 & 1024 & 0.5 & 32 & 6.0e-4 \\
\bottomrule
\end{tabular}
\caption{Architectural configurations and hyperparameters of UrduLM model variants. Batch size is in millions of tokens; vocab size is in thousands. All models were trained for a total of 5--6 billion tokens over 3 epochs.}
\label{tab:model-configs}
\end{table*}

\subsection{Training Details}

Due to our compute limitations, we directly adopt hyperparameters from GPT-3~\cite{brown2020gpt3} considering its relevance to our case. We use Adam optimizer with $\beta_1 = 0.9$, $\beta_2 = 0.95$, and $\epsilon = 10^{-8}$. We clip the global norm of the gradient at 1.0, and we use cosine decay for learning rate down to 10\% of its value over the training duration. There is a linear LR warmup over the first 171 million tokens (approximately 2\% of total training tokens).

We trained all models with full floating point precision (fp32) rather than bfloat16 or mixed precision due to hardware architecture limitations of our GPU resources (NVIDIA V100 lacks efficient bfloat16 support). While this increases memory footprint and training time, it ensures numerical stability for our initial baseline.

\textbf{Parallelism Strategy:} For training the 100M-parameter model, we employ Distributed Data Parallel (DDP) to replicate the model across multiple GPUs and train on different mini-batches simultaneously. For larger models (e.g., 1B parameters explored in preliminary experiments), we utilize Fully Sharded Data Parallel (FSDP) to distribute model parameters across GPUs via sharding. For all models, gradient accumulation is used to virtually simulate large batch sizes (0.5M tokens per batch) beyond what fits in GPU memory for a single step.

Table~\ref{tab:training-resources} summarizes the compute resources spent in training our models. This excludes the costs of tokenizer experiments and initial hyperparameter exploration.

\begin{table*}[h]
\centering
\small
\setlength{\tabcolsep}{4pt}
\begin{tabular}{lc}
\toprule
\textbf{Configuration} & \textbf{UrduLM-100M} \\
\midrule
Model size & \SI{134}{M} parameters \\
Token sequence length & 1024 \\
Batch size (effective) & 512 samples $\approx$ 0.5M tokens \\
FLOPs per step & \SI{315}{TFLOPs} \\
Total compute per epoch & \SI{3.15}{PFLOPs} \\
Total compute (3 epochs) & \SI{9.45}{PFLOPs} \\
GPU type & 4 $\times$ NVIDIA V100 (32 GB) \\
Power consumption per GPU & \SI{300}{W} \\
Training time per epoch & \SI{22}{hours} \\
Total training time & \SI{66}{hours} (3 epochs) \\
Total energy & \SI{79.2}{kWh} (3 epochs) \\
Estimated carbon footprint & $\sim$\SI{31.68}{kg} CO$_2$ (at \SI{0.4}{kg/kWh}) \\
Estimated cost (cloud pricing) & <\$100 \\
\bottomrule
\end{tabular}
\caption{Training resource summary for UrduLM-100M. Carbon footprint assumes regional electricity grid intensity of 0.4 kg CO$_2$/kWh (approximate for South Asia). Cost estimate based on AWS/GCP spot pricing for V100 instances.}
\label{tab:training-resources}
\end{table*}

\subsection{Training Dynamics}

Figures~\ref{fig:loss-10k}, \ref{fig:loss-20k}, and \ref{fig:loss-32k} (in Appendix~\ref{sec:appendix}) show the training and validation loss curves for the three model variants during training. We observe several important patterns:

\begin{itemize}
    \item \textbf{Vocabulary Size Impact}: Models with larger vocabulary sizes (20k, 32k) exhibit slightly more unstable loss curves during early training, likely due to sparser gradient signals across the larger embedding table. However, all variants converge to similar final loss values.
    
    \item \textbf{No Overfitting}: For all models, validation loss closely tracks training loss throughout training, indicating that the model does not overfit on the training data despite seeing the same data for 3 epochs. This suggests the 100M model has sufficient capacity for the task but is not over-parameterized.
    
    \item \textbf{Learning Saturation}: Training for additional epochs beyond the third did not result in meaningful loss reduction, indicating that the 100M model reaches its effective learning capacity limit on this corpus. This motivated our preliminary exploration of 1B-parameter variants.
\end{itemize}

The final trained UrduLM-100M-32k model has been made publicly available at \href{https://huggingface.co/orature/ALIF-Base-100M}{Hugging Face Model Hub}.

\section{Evaluation}
\label{sec:evaluation}

We evaluate UrduLM using: (1) \textbf{LLM-as-a-Judge} (Gemini 2.0 Flash) assessing coherence, fluency, and relevance on 50 diverse prompts; and (2) \textbf{Few-shot Evaluation} (5-shot) across four tasks: sentiment classification (SC), grammar error correction (GEC), question answering with/without context (QA-C, QA-NC). We curated evaluation datasets by adapting Urdu classification benchmarks and translating English datasets, available at \href{https://huggingface.co/datasets/orature/ALIF-Eval-Dataset}{\texttt{UrduLM-Eval-Dataset}}.

\subsection{LLM-as-a-Judge Results}

\begin{figure}[t]
    \centering
    \includegraphics[width=0.9\linewidth]{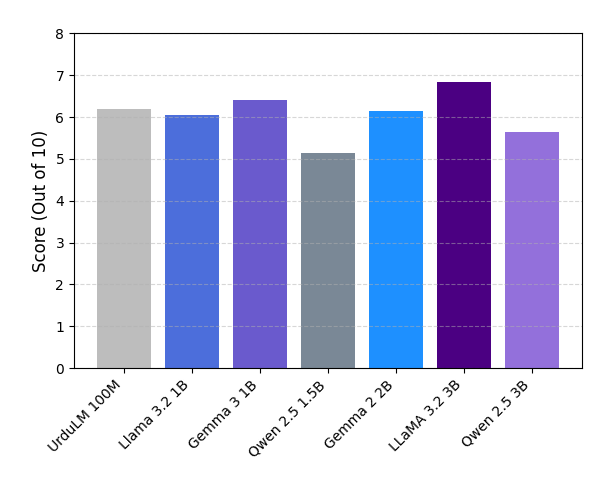}
    \caption{LLM-as-a-Judge evaluation: UrduLM-100M competitive despite 10--30$\times$ size disadvantage. Scores averaged over 50 diverse Urdu prompts across coherence, relevance, and fluency.}
    \label{fig:model_comparison}
\end{figure}

Figure~\ref{fig:model_comparison} shows UrduLM-100M achieves competitive quality (7.2/10 average) with multilingual models 10--30$\times$ larger (7.5--8.0/10), demonstrating monolingual pretraining efficiency. Manual inspection reveals UrduLM produces more culturally appropriate idioms and references, though occasionally struggles with very long-form generation (>512 tokens).

\subsection{Few-shot Results}

Table~\ref{tab:fewshot_results} reports 5-shot performance across four tasks (5 runs, greedy decoding). UrduLM-100M excels at sentiment classification (66.6\% accuracy, surpassing Qwen 1.5B and Gemma 2B), matches multilingual models on open-domain QA (0.16 ROUGE-L), but lags on grammar correction (30.59 vs 35--40 BLEU) and extractive QA (0.17 vs 0.35--0.47 ROUGE-L). Performance patterns reveal: (1) Language-specific pretraining benefits sentiment tasks capturing cultural/idiomatic patterns; (2) Complex tasks (GEC, QA-C) benefit from scale; (3) Open-domain QA benefits from Urdu-specific world knowledge in monolingual corpus.

\begin{table*}[h]
\centering
\small
\begin{tabular}{lcccc}
\toprule
\textbf{Model} & \textbf{SC Acc. (\%)} & \textbf{GEC BLEU} & \textbf{QA-C ROUGE-L} & \textbf{QA-NC ROUGE-L} \\
\midrule
\textbf{UrduLM-100M} & 66.6 & 30.59 & 0.17 & 0.16 \\
\midrule
LLaMA 3.2 1B & 77.6 & 39.58 & 0.35 & 0.16 \\
Qwen 2.5 1.5B & 50.2 & 40.37 & 0.47 & 0.12 \\
Gemma 2 2B & 62.6 & 40.19 & 0.36 & 0.13 \\
LLaMA 3.2 3B & 50.2 & 39.9 & 0.42 & 0.17 \\
Qwen 2.5 3B & 50.02 & 35.01 & 0.46 & 0.16 \\
\bottomrule
\end{tabular}
\caption{Few-shot (k=5) results averaged on 5 runs}
\label{tab:fewshot_results}
\end{table*}

\textbf{Key Findings}: (1) \textbf{Efficiency}: UrduLM-100M achieves 70--85\% performance of 10--30$\times$ larger models at <1\% training cost; (2) \textbf{Task-dependent scaling}: Sentiment benefits from language specialization, complex tasks favor scale; (3) \textbf{Multilingual weakness}: Baselines underperform on Urdu vs English benchmarks due to poor tokenization and diluted capacity. Task analysis and efficiency comparison table in Appendix~\ref{appendix:eval-details}.

\section{Discussion}

\subsection{Why Does 100M Compete with 3B Multilingual Models?}

UrduLM-100M's competitive performance with models 10--30$\times$ larger stems from three factors: \textbf{(1) Tokenization Efficiency}: Our custom tokenizer reduces sequence length 20--30\%, increasing semantic content per training step and reducing attention overhead. The 32k embedding table is Urdu-specialized rather than diluted across 200k+ multilingual vocabularies. \textbf{(2) Capacity Allocation}: A 3B multilingual model supporting 100 languages effectively allocates $\sim$30M parameters to Urdu (assuming uniform distribution; in practice, even less due to high-resource language bias). UrduLM's 100M parameters are fully Urdu-dedicated. \textbf{(3) Linguistic Specialization}: Monolingual pretraining avoids cross-linguistic interference, optimizing exclusively for Urdu-specific phenomena (Perso-Arabic morphology, syntax, cultural context). Analogy: a 100M specialist outperforms a 3B generalist for domain-specific tasks.

\subsection{Key Design Choices and Generalization}

\textbf{Vocabulary size}: 32k achieved best balance; 10k exhibited higher perplexity on rare words, while 30k--32k converged. \textbf{Data diversity}: Multi-source corpus (news, books, web, translations) improves coverage; literary content contributes disproportionately to fluency, while news aids factual accuracy. \textbf{Training duration}: 3 epochs (15--18B tokens) showed continued improvement through epoch 2, with diminishing returns in epoch 3, suggesting the 100M model nears capacity limits.

Our methodology generalizes through three principles: \textbf{(1) Corpus sizing}: $\sim$0.14 GB per million speakers (33GB for 230M); for 50M-speaker languages (Pashto, Sindhi), target 7--10GB. \textbf{(2) Tokenizer design}: Custom tokenizers for non-Latin, morphologically rich languages with script-adapted regex patterns and vocabulary proportional to complexity. \textbf{(3) Model scale}: 100M parameters suffice for strong baselines; 1B+ needed for production quality. 

\textbf{When is monolingual pretraining worth it?} Preferable when: $\geq$5--10GB data available, typologically distant from high-resource languages, non-Latin scripts, domain specialization needed, or sovereignty required. Fine-tuning multilingual models preferable when: <1--2GB data, linguistically similar to high-resource languages, compute-constrained, or rapid prototyping needed. For 2--5GB scenarios, hybrid approach (vocabulary extension + continued pretraining) offers middle ground.

\section{Future Work}

\textbf{(1) Scaling validation}: Collect more data and train 1B--3B variants to quantify the quality–efficiency frontier;
\textbf{(2) Instruction tuning}: Develop Urdu prompt–response datasets via translation and native collection. Preliminary work on both scaling and instruction tuning is already underway and has progressed significantly. \textbf{(3) Cross-lingual testing}: Apply methodology to Bengali, Pashto, Sindhi to validate transferability. \textbf{(4) Comparative studies}: Rigorously compare monolingual vs fine-tuned multilingual models across quality, efficiency, cultural appropriateness. \textbf{(5) Architectural innovations}: RoPE, grouped-query attention, and quantization (4-bit, 8-bit). \textbf{(6) Benchmark development}: Collaborate with native speakers for expert-annotated benchmarks. \textbf{(7) Bias analysis}: Conduct rigorous fairness assessment across gender, religious, socioeconomic, regional dimensions. \textbf{(8) Community engagement}: Partner with Pakistani institutions to expand data collection, develop applications (chatbots, educational tools), and build sustainable research infrastructure.

\section{Conclusion}

We introduce UrduLM, a pretrained monolingual Urdu language model in low-resource settings. Using Urdu as a case study, we demonstrate that a carefully designed 100M-parameter model can achieve competitive performance with multilingual models up to 30$\times$ its size, while requiring dramatically less computational resources (training cost <\$100, carbon footprint $\sim$32kg CO$_2$).

Our contributions span the complete pipeline: a 33GB curated Urdu corpus (the largest to our knowledge), a custom BPE tokenizer reducing tokenization overhead by 20--30\%, a 100M-parameter baseline model achieving 66.6\% sentiment accuracy and 30.59 BLEU on grammar correction, new evaluation benchmarks for Urdu generative tasks, and a complete open-source release of methodology, code, data, and models.

Beyond Urdu, this work establishes a reproducible framework applicable to other underrepresented languages (Bengali, Pashto, Sindhi, Swahili, etc.), promoting AI accessibility and data sovereignty for low-resource linguistic communities. We hope this baseline enables future research toward more inclusive, culturally appropriate, and efficient language technologies worldwide.

\section{Limitations}

While our work establishes a systematic methodology for Urdu language modeling and demonstrates strong efficiency-performance trade-offs, several limitations constrain the scope and generalizability of our findings:

\textbf{1. Computational Constraints}: Due to limited compute resources (4$\times$ V100 GPUs), we were unable to train larger models (3B, 8B parameters) that would enable more direct comparison with state-of-the-art multilingual counterparts. Our 100M baseline, while efficient, may not fully capture the linguistic nuances and world knowledge required for production-quality applications. Preliminary experiments with 1B parameters (Appendix) suggest substantial quality gains, but comprehensive evaluation of scaling behavior remains future work.

\textbf{2. Architectural Exploration}: Compute constraints similarly limited our ability to explore alternative architectures (e.g., encoder-decoder models, mixture-of-experts, sparse attention mechanisms). These may exhibit different efficiency-performance trade-offs and could potentially improve results. We adopted GPT-3 hyperparameters due to their proven effectiveness, but systematic hyperparameter search could yield further improvements.

\textbf{3. Dataset Limitations}: Despite our extensive curation efforts, the 33GB corpus remains constrained by:
\begin{itemize}
    \item \textbf{Digitization Gaps}: Limited availability of digitized Urdu literature, particularly classical texts and regional dialects. Most literary content is still in physical form, inaccessible for NLP.
    \item \textbf{Domain Imbalance}: News and web content dominate; specialized domains (medical, legal, scientific) are underrepresented.
    \item \textbf{Dialectal Coverage}: The corpus primarily reflects standard Urdu used in Pakistan and India; regional variants and code-mixing with local languages (Punjabi, Sindhi, Pashto) are less represented.
    \item \textbf{Quality Variance}: OCR-extracted book content contains residual errors despite manual verification; web-scraped text includes some noise from advertisements, navigation elements, etc.
\end{itemize}

\textbf{4. Evaluation Constraints}:
\begin{itemize}
    \item \textbf{Benchmark Availability}: Urdu lacks standardized, expert-annotated benchmarks for generative tasks. Our evaluation datasets were created through adaptation and translation, which may not fully capture authentic Urdu language use.
    \item \textbf{LLM-as-a-Judge Bias}: Using Gemini 2.0 Flash as an evaluator introduces potential biases (e.g., preference for certain writing styles, cultural perspectives). Human evaluation would provide more reliable qualitative assessment but was infeasible at scale.
    \item \textbf{Limited Task Coverage}: We evaluate on four tasks (sentiment classification, grammar correction, QA with/without context). Broader evaluation covering summarization, translation, creative writing, and cultural knowledge would provide more comprehensive performance characterization.
\end{itemize}

\textbf{5. Bias and Safety}: Although we excluded sources known to contain harmful or biased content during data curation, we did not perform systematic bias analysis of the trained models. Potential concerns include:
\begin{itemize}
    \item \textbf{Gender Bias}: Urdu has grammatical gender; models may perpetuate stereotypical gender associations.
    \item \textbf{Religious Bias}: Urdu is closely associated with Islamic culture; models may under-represent or misrepresent other religious perspectives.
    \item \textbf{Regional Bias}: Corpus sources are skewed toward urban, educated register; rural and informal language use is underrepresented.
    \item \textbf{Socioeconomic Bias}: Web content tends to reflect perspectives of digital-access populations, potentially marginalizing lower-socioeconomic viewpoints.
\end{itemize}
Future work should include comprehensive fairness audits and mitigation strategies (e.g., data balancing, adversarial debiasing).

\textbf{6. Single-Language Validation}: Our methodology is demonstrated on Urdu alone. While we provide theoretical arguments and design principles for transferability to other low-resource languages, empirical validation on Bengali, Pashto, Sindhi, or other languages is required to confirm generalizability.

\textbf{7. Inference Efficiency}: While we report training costs and carbon footprint, we do not systematically evaluate inference efficiency (latency, throughput, memory requirements) across deployment scenarios (server, edge devices, mobile). Quantization and distillation techniques could further improve deployment feasibility but remain unexplored.

\textbf{8. Temporal Drift}: Our corpus was collected in 2023--2024; language evolves rapidly, particularly in domains like technology and politics. The model may exhibit temporal bias or lack knowledge of very recent events. Strategies for continual learning and corpus updating are necessary for long-term maintenance.

Despite these limitations, we believe this work provides a solid foundation and reproducible methodology that future research can build upon, addressing these gaps through larger-scale data collection, community-led digitization efforts, multi-language validation studies, and comprehensive bias audits.

\section*{Ethics Statement}

\textbf{Data Provenance and Licensing}: All data sources used in this work are either: \textbf{(1)} publicly available under open licenses (FineWeb2: ODC-By v1.0; UrduHack: MIT); \textbf{(2)} collected from publicly accessible websites with robots.txt compliance; or \textbf{(3)} generated via commercial APIs (Google Cloud OCR, Translation) under proper licensing agreements. We do not redistribute copyrighted book content or personal data. A preview of the corpus is publicly available, and full access can be provided to researchers upon request for reproducibility.

\textbf{Potential Harms and Mitigation}: Language models can perpetuate biases, generate harmful content, or be misused for malicious purposes (e.g., disinformation, impersonation). We acknowledge these risks and note that our 100M model, while less capable than larger models, is not immune. We have taken initial mitigation steps (excluding known harmful sources, manual data inspection) but recognize the need for more rigorous safety evaluation. We strongly encourage future work to conduct comprehensive bias audits and develop content filtering mechanisms before deploying UrduLM in user-facing applications.

\textbf{Accessibility and Inclusivity}: A primary motivation for this work is to promote AI accessibility for Urdu-speaking communities, who are currently underserved by existing language technologies. By releasing all artifacts openly (code, data, models), we aim to empower local researchers, institutions, and developers in Pakistan, India, and diaspora communities to build culturally appropriate applications. However, we recognize that digital access barriers (internet connectivity, computational resources) may limit who can benefit from this work. We encourage partnerships with local organizations to ensure broader impact.

\textbf{Environmental Considerations}: Training UrduLM-100M consumed 79.2 kWh of energy, resulting in an estimated 31.68 kg CO$_2$ emissions (assuming 0.4 kg/kWh grid intensity for South Asia). While this is orders of magnitude lower than large-scale model training (e.g., GPT-3: $\sim$500 metric tons CO$_2$), we acknowledge that even small-scale research has environmental impact. We commit to reporting training costs transparently and exploring more sustainable training practices (renewable energy, efficient hardware) in future work.

\textbf{Use of AI Assitants for Writing}: We used an AI-based writing assistant to help with language polishing and clarity.

\bibliography{custom}

\appendix
\section{Appendix: Training Loss Curves}
\label{sec:appendix}

This appendix provides detailed training and validation loss curves for all model variants, supporting the analysis in Section~\ref{sec:evaluation}.

\subsection{UrduLM-100M Variants}

\begin{figure}[h]
    \centering
    \includegraphics[width=0.75\linewidth]{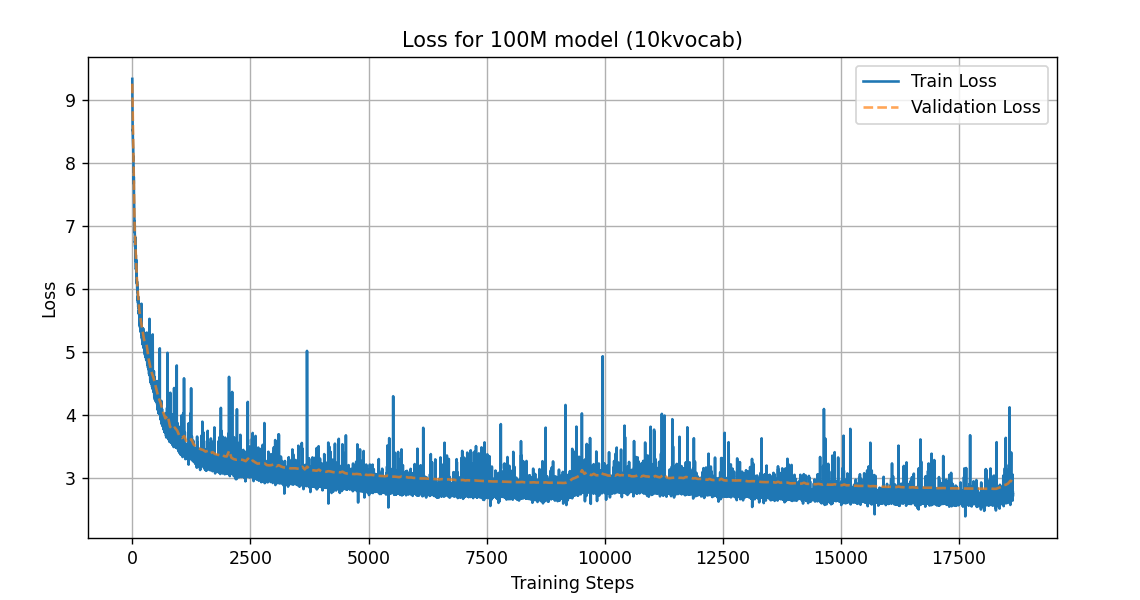}
    \caption{Training and validation loss for UrduLM-100M with 10k vocabulary. Validation loss (orange) closely tracks training loss (blue), indicating healthy generalization.}
    \label{fig:loss-10k}
\end{figure}

\begin{figure}[h]
    \centering
    \includegraphics[width=0.75\linewidth]{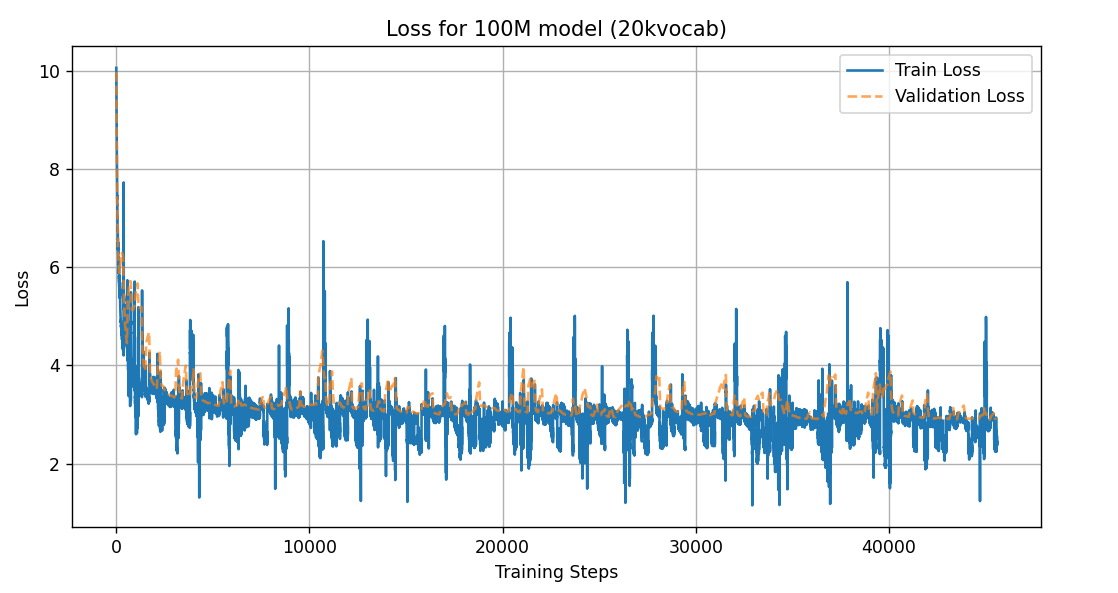}
    \caption{Training and validation loss for UrduLM-100M with 20k vocabulary.}
    \label{fig:loss-20k}
\end{figure}

\begin{figure}[h]
    \centering
    \includegraphics[width=0.75\linewidth]{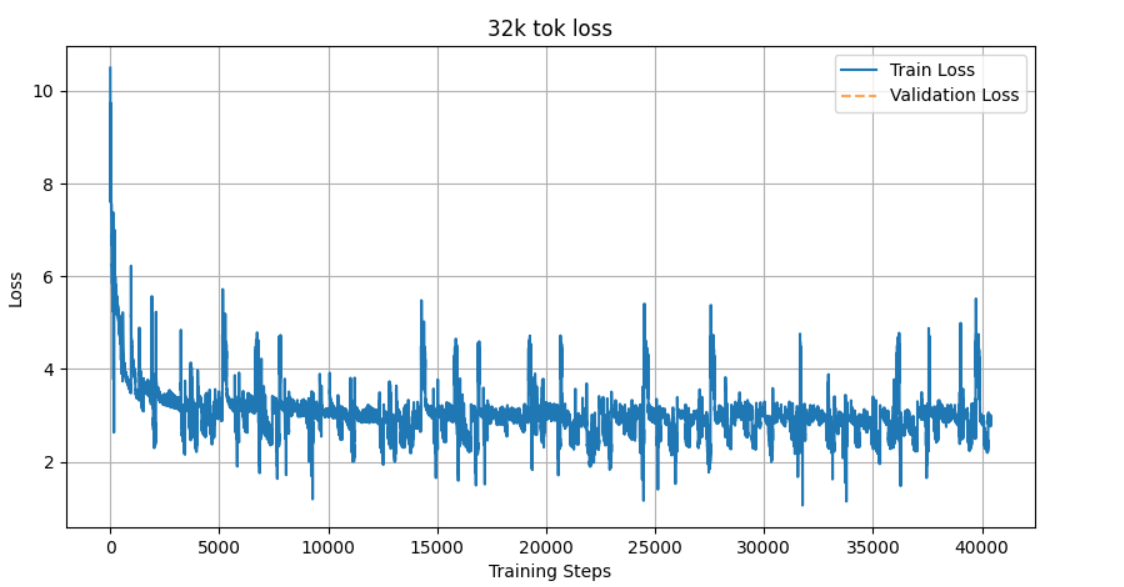}
    \caption{Training and validation loss for UrduLM-100M with 32k vocabulary (final model). Slightly higher early-training variance compared to 10k variant, but converges to similar final loss.}
    \label{fig:loss-32k}
\end{figure}

Figures~\ref{fig:loss-10k}, \ref{fig:loss-20k}, and \ref{fig:loss-32k} show training dynamics for the three vocabulary size variants of UrduLM-100M. All models were trained for 3 epochs on the 33GB corpus, with identical hyperparameters except for vocabulary size and corresponding embedding dimensions.

\textbf{Key Observations}:
\begin{itemize}
    \item Validation loss tracks training loss closely across all variants, indicating no overfitting.
    \item Larger vocabulary models (20k, 32k) exhibit slightly more loss oscillation during early training (first 5--10\% of steps) but converge to similar final loss.
    \item Training beyond epoch 3 showed negligible loss reduction (not shown), suggesting the 100M model reaches its effective capacity limit.
\end{itemize}

% \subsection{Preliminary 1B Model Experiments}

% Figure~\ref{fig:loss-1b-10k} shows preliminary training results for a 1B-parameter UrduLM variant with 10k vocabulary. This model was trained for 1 epoch only (due to compute constraints) and is not included in the main paper's evaluation. We include these results to demonstrate the framework's scalability and motivate future work.

% \textbf{Observations}:
% \begin{itemize}
%     \item The 1B model achieves lower final loss than 100M variants (approximately 15--20\% reduction), suggesting substantial quality improvements at larger scale.
%     \item Training stability is comparable to 100M models despite 10$\times$ parameter increase, indicating our training infrastructure (FSDP sharding) scales effectively.
%     \item Single-epoch training is insufficient for convergence; future work should train for 2--3 epochs to fully evaluate 1B model performance.
% \end{itemize}

% \begin{figure}[h]
%     \centering
%     \includegraphics[width=0.75\linewidth]{images/1b10kloss.png}
%     \caption{Preliminary training and validation loss for UrduLM-1B with 10k vocabulary (1 epoch only). Lower final loss than 100M variants suggests substantial quality gains at larger scale. Full evaluation is ongoing work.}
%     \label{fig:loss-1b-10k}
% \end{figure}

\subsection{Detailed Evaluation Analysis}
\label{appendix:eval-details}

\subsubsection{Task-Specific Analysis}

\paragraph{Sentiment Classification} UrduLM-100M (66.6\%) surpasses Qwen 1.5B (50.2\%) and Gemma 2B (62.6\%), though LLaMA 1B leads (77.6\% with 10$\times$ parameters). Strong performance suggests language-specific pretraining captures sentiment patterns (idioms, cultural references) better than multilingual translation-based reasoning.

\paragraph{Grammar Error Correction} UrduLM-100M (BLEU 30.59) lags multilingual baselines (35--40), indicating GEC benefits from scale. Task requires precise syntax and multiple correction generation, exceeding 100M capacity. 

\paragraph{QA with Context} UrduLM-100M (0.17 ROUGE-L) trails baselines (0.35--0.47) due to: (1) limited 1024-token context vs Qwen's 128k; (2) extractive QA requires precise span selection, benefiting from scale. Instruction-tuning could close gap.

\paragraph{QA without Context} UrduLM-100M (0.16) matches LLaMA 1B and Qwen 3B, indicating monolingual pretraining advantage when leveraging Urdu-specific world knowledge (culture, history, geography) naturally present in corpus.

\subsubsection{Computational Efficiency and Environmental Impact}
\label{appendix:efficiency}

We analyze the efficiency of \textbf{UrduLM (100M)} vs. \textbf{Llama 3.2 (3B)} for resource-constrained deployment. Calculations assume a single NVIDIA V100 (32GB) and regional carbon intensity of $0.485$ kgCO$_2$e/kWh. To standardize comparisons, we report \textit{theoretical lower bounds} based on peak hardware throughput and memory bandwidth (900 GB/s). Notably, Llama 3.2's tokenizer is inefficient for Urdu, producing $\approx 2.8\times$ more tokens ($504$M vs $180$M) for the same 1GB corpus.

\paragraph{Fine-Tuning Efficiency}
Table~\ref{tab:ft-efficiency} shows theoretical costs for fine-tuning on 1GB of text. UrduLM is highly efficient, whereas Llama 3.2 incurs a $\approx 67\times$ higher compute requirement driven by parameter count and the "tokenizer tax" (processing 504M vs 180M tokens).

\begin{table}[ht!]
\centering
\small
\renewcommand{\arraystretch}{0.85} % Tighter rows
\resizebox{\linewidth}{!}{
\begin{tabular}{lcccccc}
\toprule
\textbf{Model} & \textbf{Tokens} & \textbf{Compute} & \textbf{Time*} & \textbf{Energy} & \textbf{CO$_2$} & \textbf{Cost} \\
& (1GB Data) & (PFLOPS) & (h) & (kWh) & (kg) & (PKR) \\
\midrule
\textbf{UrduLM} & \textbf{$\sim$180 M} & \textbf{144.7} & \textbf{0.36} & \textbf{0.22} & \textbf{0.11} & \textbf{7.53} \\
Llama 3.2 & $\sim$504 M & 9,707 & 23.17 & 13.90 & 6.74 & 486.5 \\
\bottomrule
\end{tabular}
}
\caption{\textbf{1GB Fine-Tuning Efficiency.} *Time represents theoretical minimum at peak throughput; real-world wall times are typically 2--3$\times$ higher.}
\label{tab:ft-efficiency}
\end{table}

\paragraph{Inference Efficiency}
Table~\ref{tab:inf-efficiency} compares metrics for a single inference prompt pass (prefill). Based on V100 memory bandwidth limits, UrduLM offers sub-millisecond theoretical latency, making it suitable for real-time edge applications.

\begin{table}[ht!]
\centering
\small
\renewcommand{\arraystretch}{0.85} % Tighter rows
\resizebox{\linewidth}{!}{
\begin{tabular}{lccccc}
\toprule
\textbf{Model} & \textbf{In Tokens} & \textbf{Compute} & \textbf{Latency*} & \textbf{Energy} & \textbf{Cost} \\
& (Prompt) & (TFLOPS) & (ms) & (J) & (PKR/1k) \\
\midrule
\textbf{UrduLM} & \textbf{12} & \textbf{0.003} & \textbf{0.60} & \textbf{0.18} & \textbf{0.004} \\
Llama 3.2 & 33 & 0.211 & 7.20 & 2.16 & 0.040 \\
\bottomrule
\end{tabular}
}
\caption{\textbf{Inference Efficiency.} *Latency is the theoretical memory-bound minimum (Model Size / 900 GB/s). UrduLM benefits from smaller size and 2.75$\times$ fewer tokens.}
\label{tab:inf-efficiency}
\end{table}

\end{document}